\documentclass[runningheads]{llncs}

\usepackage{eccv}

\usepackage{eccvabbrv}

\usepackage{graphicx}
\usepackage{booktabs}
\usepackage{mathrsfs}

\usepackage[accsupp]{axessibility}  
\usepackage{makecell}
\usepackage{booktabs, pifont, xcolor,bm}
\definecolor{darkgreen}{rgb}{0.0, 0.5, 0.0}
\definecolor{darkred}{rgb}{0.6, 0.0, 0.0}

\definecolor{red}{rgb}{1,0.2,0.2}
\definecolor{or}{rgb}{1,0.5,0.25}
\definecolor{green}{rgb}{0, 1, 0}
\definecolor{bl}{rgb}{0, 0, 1}
\definecolor{brown}{rgb}{0.59, 0.3, 0}
\definecolor{cyan}{rgb}{0, 1, 1}
\definecolor{c_lowbest}{rgb}{1.0,1.0,0.9}
\definecolor{c_highbest}{rgb}{0.9,1.0,0.9}

\newcommand{\best}[1]  {{\textbf{#1}}}
\newcommand{\second}[1]  {{#1}}

\usepackage{hyperref}

\usepackage{orcidlink}

\usepackage{multirow}

\begin{document}

\title{AirSplat: Alignment and Rating for Robust Feed-Forward 3D Gaussian Splatting}

\titlerunning{AirSplat}

\author{Minh-Quan Viet Bui\thanks{Co-first authors (equal contribution).}\orcidlink{0000-0001-8511-7731} \and
Jaeho Moon\protect\footnotemark[1]\orcidlink{0009-0009-6213-7700} \and
Munchurl Kim\orcidlink{0000-0003-0146-5419}}

\authorrunning{Bui et al.}

\institute{KAIST, Republic of Korea\\
\email{\{bvmquan, jaeho.moon, mkimee\}@kaist.ac.kr}\\
\small{\url{https://kaist-viclab.github.io/airsplat-site}}}

\maketitle


\begin{figure*}[!h]
    \centering
    \includegraphics[width=\linewidth]{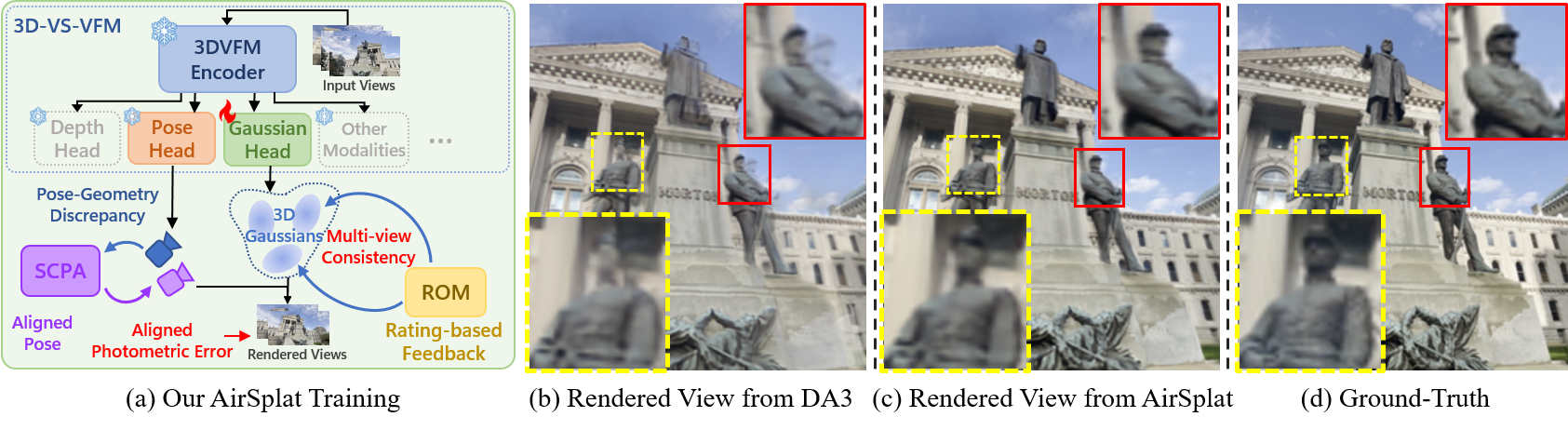}
    \vspace{-5mm}
    \caption{(a) Our proposed AirSplat adapts 3D-VS-VFMs using Self-Consistent Pose Alignment (SCPA) and Rating-based Opacity Matching (ROM) to resolve inherent pose-geometry discrepancies and multi-view inconsistencies. Compared to (b) baseline DA3 \cite{lin2025da3} which shows floaters (red boxes) and blurred regions (dashed yellow boxes), (c) AirSplat eliminates these artifacts, faithfully recovering the scene structure shown in (d) the ground truth.}
    \vspace{-3mm}
    \label{fig:teaser}
\end{figure*}


\begin{abstract}
  While 3D Vision Foundation Models (3DVFMs) have demonstrated remarkable zero-shot capabilities in visual geometry estimation, their direct application to generalizable novel view synthesis (NVS) remains challenging.
  In this paper, we propose \textbf{AirSplat}, a novel training framework that effectively adapts the robust geometric priors of 3DVFMs into high-fidelity, pose-free NVS. Our approach introduces two key technical contributions: 
  (1) \textit{Self-Consistent Pose Alignment} (SCPA), a training-time feedback loop that ensures pixel-aligned supervision to resolve pose-geometry discrepancy; and 
  (2) \textit{Rating-based Opacity Matching} (ROM), which leverages the local 3D geometry consistency knowledge from a sparse-view NVS teacher model to filter out degraded primitives.
  Experimental results on large-scale benchmarks demonstrate that our method \textit{significantly} outperforms state-of-the-art pose-free NVS approaches in reconstruction quality.
  Our AirSplat highlights the potential of adapting 3DVFMs to enable simultaneous visual geometry estimation and high-quality view synthesis.
\end{abstract}
\section{Introduction}
\label{sec:introduction}

The field of Novel View Synthesis (NVS) has undergone a paradigm shift with the advent of 3D Gaussian Splatting (3DGS) \cite{kerbl20233d} and its subsequent advancements \cite{Wu_2024_CVPR, park2024splinegs, lin2025longsplat, bui2025mobgsmotiondeblurringdynamic, fu2024colmap, yu2024mip}. 
To overcome the bottleneck of time-intensive per-scene optimization, recent feed-forward architectures \cite{charatan2024pixelsplat, chen2024mvsplat, xu2025depthsplat, ziwen2025long} directly predict 3D scene parameters from sparse context views. 
However, dependence on calibrated poses limits `in-the-wild' applicability, while several pose-free feed-forward methods \cite{hong2023unifying, smart2024splatt3r,kang2025selfsplat, zhang2025flarefeedforwardgeometryappearance} remain fundamentally constrained to sparse-view inputs. 
To address this issue, recent works \cite{ye2024no, huang2025no, park2026ecosplat, jiang2025anysplat, ye2026yonosplat} leverage the robust, zero-shot depth and pose estimation capabilities of 3D Vision Foundation Models (3DVFMs) such as MASt3R \cite{mast3r_eccv24}, VGGT \cite{wang2025vggt}, and $\pi^3$ \cite{wang2025pi3}. 
These approaches fine-tune or distill 3DVFMs to jointly infer scene 3D Gaussian primitives and camera parameters directly from raw input images. 
Nevertheless, this biases the networks toward the view synthesis objective, resulting in performance degradation on foundational geometry estimation tasks. 
More recent unified frameworks, which we refer to as 3DVFMs with view synthesis (3D-VS-VFMs), such as WorldMirror \cite{liu2025worldmirror} and DepthAnything3 \cite{lin2025da3}, incorporate 3DGS heads to perform both geometry estimation and NVS. 
Despite this integration, generating high-fidelity novel views from completely uncalibrated context images remains challenging.

We identify two fundamental obstacles that inherently hinder 3D-VS-VFMs from achieving high-fidelity NVS. First, there exists a critical \textit{pose-geometry discrepancy} during the training process. 
As illustrated in Fig.~\ref{fig:pose_corr}-(a) and (b), current NVS fine-tuning strategies either fail to generalize due to a lack of direct supervision for novel viewpoints (context-only~\cite{jiang2025anysplat}) or suffer from coordinate misalignment (context-target~\cite{huang2025no}).
Specifically, in the context-target setting, target camera poses are inferred using features from both the context and target images, whereas the scene geometry is conditioned strictly on the context views to maintain feed-forward generalizability \cite{huang2025no}.
However, this asymmetric information flow induces a latent coordinate misalignment between the \textit{predicted target poses} and the \textit{context 3D Gaussian primitives}, leading to degraded optimization (Fig.~\ref{fig:pose_corr_diff}). 
Second, as shown in Fig.\ref{fig:teaser}, the renderings of current 3D-VS-VFMs often contain local multi-view inconsistencies. Specifically, corresponding primitives generated from different context views lack precise spatial consensus. This deficiency leads to local geometric inconsistencies and the generation of `floaters' that severely degrade spatial stability. 


To bridge these fundamental gaps, we introduce \textbf{AirSplat}, a pose-free feedforward 3DGS training framework driven by self-consistent pose \textbf{A}l\textbf{i}gnment and \textbf{R}ating-based feedback that adapts state-of-the-art (SOTA) 3D-VS-VFMs. 
First, we present \textbf{Self-Consistent Pose Alignment (SCPA)} to dynamically anchor the predicted target poses to the scene geometry derived from the context views, thereby providing geometrically-consistent reconstruction supervision. 
By re-estimating the camera pose from a rendered proxy image, we isolate the systematic geometric bias between the initial pose prediction and the network's 3D Gaussian primitives prediction. We then apply an inverse correction to the initial target pose, aligning it with the estimated 3D primitives, as in Fig.~\ref{fig:pose_corr}-(c). This entirely decouples the scene reconstruction from the coordinate frame drift during training. 
Next, we introduce \textbf{Rating-based Opacity Matching (ROM)}, a novel optimization strategy designed to enforce multi-view consistency among predicted primitives and eliminate geometrically inconsistent artifacts. This approach is inspired by the paradigm of \textit{learning from rating-based feedback} \cite{10.5555/3305890.3305917, arumugam2019deep, white2024rating, luu2025enhancing}, where an agent's behavior is refined using positive and negative evaluations from an external teacher. In standard rating-based frameworks, an agent is optimized to ensure that its predicted ratings accurately align with the collected feedback from a human or AI evaluator. Adapting this paradigm to 3D reconstruction learning, we utilize a lightweight, sparse-view feed-forward 3DGS model as an algorithmic `teacher oracle' to provide geometric ratings for the primitives estimated by our network. This feedback effectively partitions the predicted 3D space into preferred (geometrically consistent) and rejected (inconsistent) states. Crucially, we directly formulate a primitive's predicted rating as its opacity. By strictly matching the predicted opacity to the teacher's geometric rating, \textbf{AirSplat} implicitly prunes spatial artifacts. 
In summary, our core \textbf{contributions} are as follows:
\begin{itemize}
  \setlength\itemsep{0.1cm}
  \item We introduce \textbf{AirSplat}, a novel framework that fine-tunes 3D-VS-VFMs to generate high-fidelity novel views, while preserving the robust geometry estimation performance.
  \item We propose \textbf{Self-Consistent Pose Alignment (SCPA)} to resolve pose-geometry discrepancies, thereby preventing model degradation caused by misaligned photometric supervision.
  \item We design \textbf{Rating-based Opacity Matching (ROM)}, an optimization strategy that learns from geometric rating-based feedback from a lightweight teacher model to seamlessly filter inconsistent and floating primitives.
  \item Our AirSplat achieves the state-of-the-art NVS performance \textit{with large margins} on dense-view benchmarks, including RealEstate10K \cite{zhou2018stereo}, DL3DV \cite{ling2024dl3dv}, and ACID \cite{infinite_nature_2020}, demonstrating robustness in pose-free settings.
\end{itemize}

\section{Related Work}
\label{sec:related_work}

\subsection{Optimization-based NVS}

The paradigm of novel view synthesis (NVS) has seen a dramatic shift from Neural Radiance Fields (NeRF)~\cite{mildenhall2020nerf}. 
While several NeRF variants~\cite{yu2021plenoctrees, fridovich2022plenoxels, muller2022instant, tensorf, 10504815} successfully compressed training and inference times, 3D Gaussian Splatting (3DGS)~\cite{kerbl20233d} broke new ground by modeling scenes with anisotropic primitives and a differentiable rasterizer.
Subsequent research \cite{yu2024mip, huang20242d, fu2024colmap} has refined this representation to improve reconstruction quality and pose robustness.
To mitigate costly per-scene optimization of 3DGS, a parallel line of work seeds the optimization with stronger priors instead of a sparse SfM point cloud: InstantSplat~\cite{fan2024instantsplat} initializes the Gaussians from the dense points of a geometric foundation model in a pose-free manner, QuickSplat~\cite{liu2025quicksplat} learns data-driven priors that jointly predict the initial Gaussians and a learned densification policy, and DAPS-AGF~\cite{yousaf2025daps} introduces depth-aware supervision with adaptive gradient-based densification for weakly-observed regions.
However, even with such accelerated initialization, these methods still incur seconds of iterative, per-scene refinement for every new environment, preventing the immediate translation of raw pixels into 3D structures.
This persistent reliance on scene-specific optimization creates a significant barrier for applications requiring low-latency and scalability, underscoring the need for generalizable, feed-forward architectures.

\subsection{3D Vision Foundation Models (3DVFMs)}
\label{sec:related_works_3DVFM}
The rapid scaling of 2D vision models has recently extended into the 3D domain, giving rise to 3DVFMs capable of broad, zero-shot geometric reasoning. Unlike conventional Structure-from-Motion (SfM) pipelines \cite{schonberger2016structure, DBLP:journals/tog/SnavelySS06} which infer 3D structure through iterative and computationally expensive bundle adjustment, 3DVFMs approach geometry estimation as a feed-forward prediction task. 
DUSt3R \cite{wang2024dust3r} estimates dense corresponding point maps from unposed image pairs. 
MASt3R \cite{mast3r_eccv24} further advances this by incorporating dense feature matching heads to enhance correspondence learning. 
VGGT \cite{wang2025vggt} adopts an alternative attention mechanism to generalize across a variable number of input views. 
$\pi^3$ \cite{wang2025pi3} explores permutation-equivariant prediction to eliminate reference coordinate bias.
While the majority of 3DVFMs focus strictly on explicit geometric outputs (e.g., depth, point map, correspondence), recent variants such as WorldMirror \cite{liu2025worldmirror} and DepthAnything3 (DA3) \cite{lin2025da3} are equipped with dedicated 3DGS predictions for NVS tasks. 
We formally classify this specialized subset of architectures as 3DVFMs with a view synthesis head (3D-VS-VFMs).



\subsection{Feed-forward NVS}
Early generalizable architectures \cite{pixelnerf, mvsnerf, ibrnet, gpnr} utilized Transformer-based encoders to aggregate image features into NeRF \cite{mildenhall2020nerf} or image-based rendering models. The advent of feed-forward 3DGS \cite{charatan2024pixelsplat, szymanowicz2024splatter, chen2024mvsplat, xu2025depthsplat, ziwen2025long, wang2025zpressor} has redefined this frontier by directly mapping pixels to 3D Gaussian primitives. Despite their efficiency, these methods typically rely on calibrated camera parameters extracted from off-the-shelf SfM pipelines \cite{schonberger2016structure, DBLP:journals/tog/SnavelySS06}, which fundamentally restricts their applicability in spontaneous, unconstrained environments. 

To bypass SfM dependency, several pose-free feed-forward NVS methods \cite{smart2024splatt3r, hong2023unifying, kang2025selfsplat, zhang2025flarefeedforwardgeometryappearance} primarily focused on sparse-view reconstruction. 
More recently, a new paradigm has emerged that initializes directly from pre-trained 3DVFMs. 
Building upon MASt3R \cite{mast3r_eccv24}, NoPoSplat \cite{ye2024no} estimates 3D Gaussian primitives in canonical space to avoid explicit pose estimation noise, while SPFSplat \cite{huang2025no} proposes to jointly learn pose estimation and NVS, utilizing reprojection losses. 
AnySplat \cite{jiang2025anysplat}, building upon VGGT \cite{wang2025vggt}, jointly optimizes camera poses and 3DGS predictions through photometric supervision and geometric distillation from a 3DVFM teacher model. YoNoSplat \cite{ye2026yonosplat} explores the mix-forcing training strategy for robust joint camera pose and NVS learning initialized from \cite{wang2025pi3}.
While Rayzer \cite{jiang2025rayzer}, an orthogonal approach, predicts camera and scene representations from scratch via ray-based transformers, it lacks view-count generalization, requiring retraining for varying input numbers.

Recently, 3D-VS-VFMs\cite{liu2025worldmirror, lin2025da3} have proposed learning visual geometry estimation and NVS within a single, unified model. 
However, despite their zero-shot generalization, a discernible NVS performance gap remains between these unified 3D-VS-VFMs and specialized NVS pipelines \cite{ye2026yonosplat, huang2025no}. 
In this work, we identify the critical bottlenecks in the NVS quality of 3D-VS-VFMs: pose-geometry misalignment and multi-view inconsistency. 
To resolve these, we propose AirSplat, a 3D-VS-VFM training framework that significantly boosts high-fidelity NVS performance, while preserving the integrity of the visual geometry estimation.

\section{Proposed Method}
\label{sec:method}

\begin{figure*}[t]
    \centering
    \includegraphics[width=\linewidth]{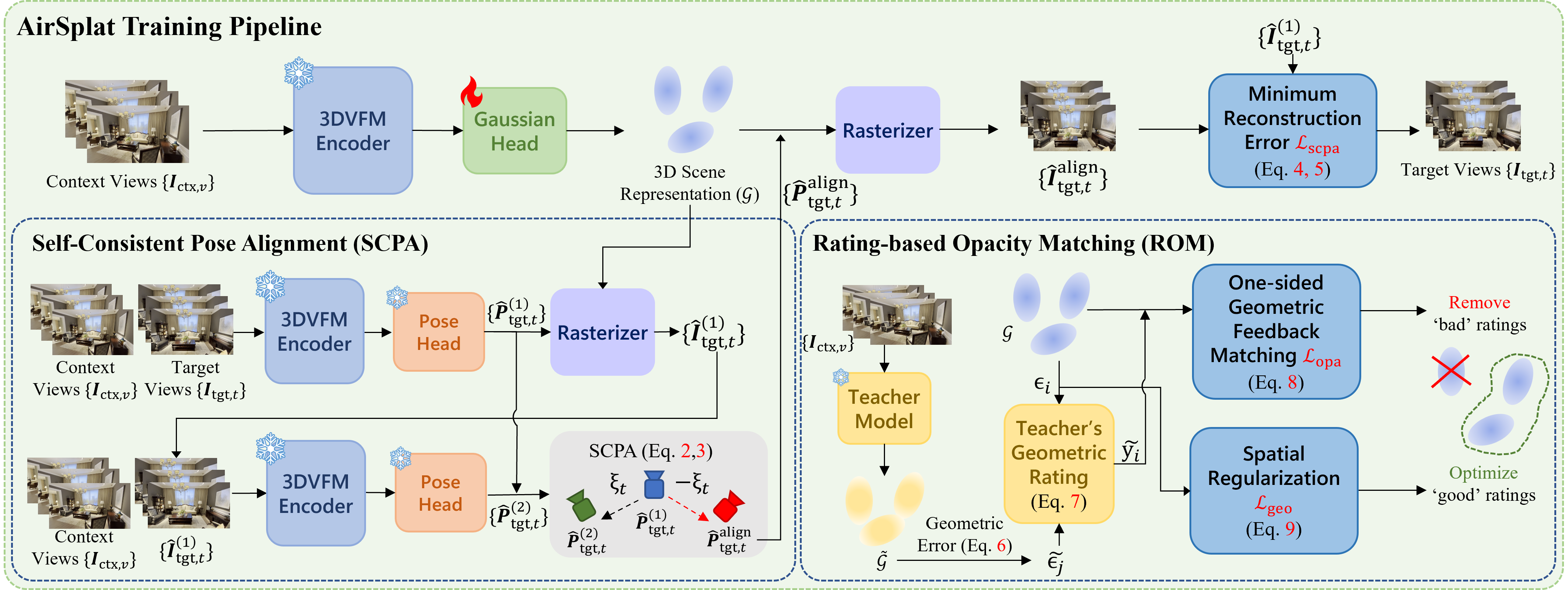}
    \caption{\textbf{Overview of our AirSplat training pipeline.} During training of a Gaussian head of a 3DVFM, the encoder and its pose head are frozen. Our SCPA module corrects the predicted target pose to align the rendered target views to the GT target views. Our ROM module gathers the geometric feedback from the teacher model, and based on the feedback, it enhances the multi-view consistency of the predicted 3D primitives.}
    \label{fig:main_architecture}
    \vspace{-3mm}
\end{figure*}

\subsection{Overview of AirSplat}
Given a set of $V$ uncalibrated input context views $\mathcal{I}_\text{ctx} = \{\bm{I}_{\text{ctx},v}\}_{v=1}^V$, our model $f_{\phi}$ predicts a set of $N$ pixel-aligned 3D Gaussian primitives $\mathcal{G} = \{g_i\}_{i=1}^N$ alongside the estimated context camera intrinsics $\hat{\mathcal{K}}_\text{ctx} = \{\hat{\bm{K}}_{\text{ctx},v}\}_{v=1}^V$ and extrinsics $\hat{\mathcal{P}}_\text{ctx} = \{\hat{\bm{P}}_{\text{ctx},v}\}_{v=1}^V$. Each individual Gaussian primitive $g_i$ is explicitly parameterized by its 3D mean position $\bm{\mu}_i \in \mathbb{R}^3$, covariance matrix $\bm{\Sigma}_i$, opacity $\alpha_i \in [0, 1]$, and color $\bm{c}_i$. 
To adapt $f_{\phi}$ for high-fidelity NVS while preserving its foundational priors, we freeze the main 3DVFM encoder and geometry heads, optimizing only the Gaussian prediction head. Our overall training framework is then driven by two core modules: Self-Consistent Pose Alignment (SCPA), which dynamically resolves pose-geometry discrepancies during target view rendering, and Rating-based Opacity Matching (ROM), which systematically prunes hallucinated artifacts using geometric feedback from a teacher model.

\begin{figure*}[t]
    \centering
    \includegraphics[width=0.9\linewidth]{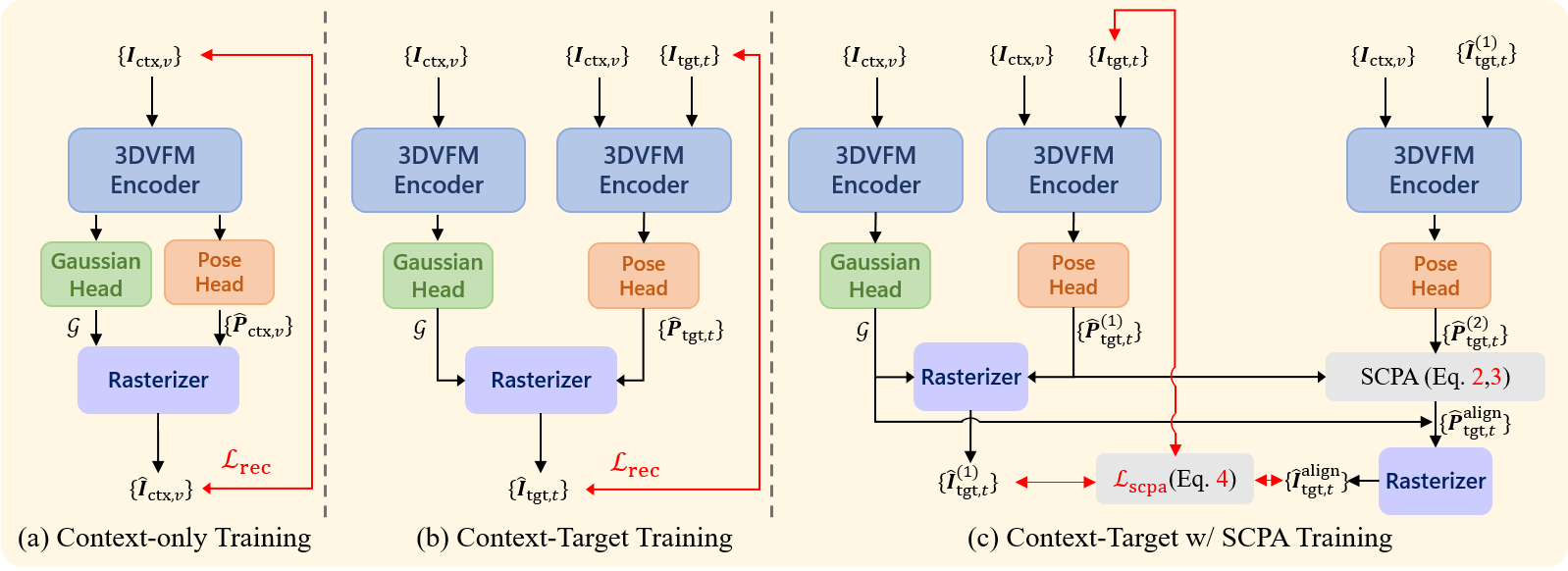}
    \vspace{-1mm}
    \caption{\textbf{Comparison of training paradigms in pose-free NVS.} (a) Training with the context-only strategy, adopted in \cite{jiang2025anysplat}, leads to a lack of direct supervision for novel viewpoints. (b) The context-target strategy, following \cite{huang2025no}, results in spatial misalignment. (c) Our SCPA corrects the inherent spatial drift, enabling the network to learn both structurally consistent 3D geometry and robust novel view synthesis.}
    \label{fig:pose_corr}
    \vspace{-5mm}
\end{figure*}

\subsection{Self-Consistent Pose Alignment (SCPA)}
\label{sec:scpa}

\begin{figure*}[t]
    \centering
    \includegraphics[width=0.95\linewidth]{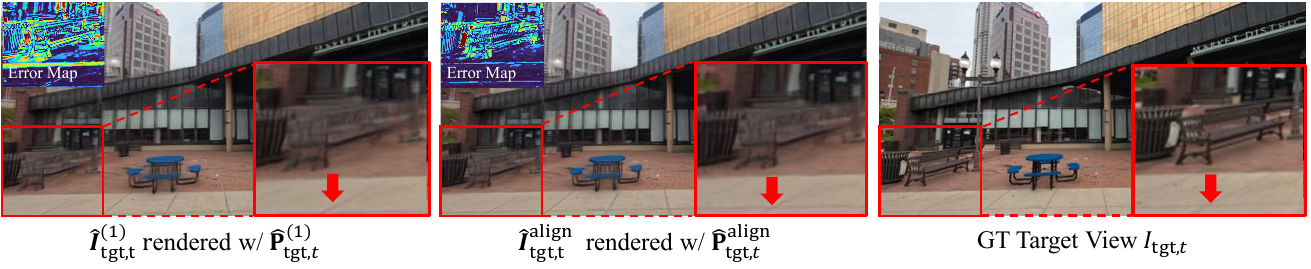}
    \caption{\textbf{Effect of Self-Consistent Pose Alignment (SCPA).} We compare rendered target views using the initial predicted pose $\hat{\bm{P}}^{(1)}_{\text{tgt},t}$, our aligned pose $\hat{\bm{P}}_{\text{tgt},t}^\text{align}$ against the ground truth, during training. As highlighted by the red arrows, the initial pose prediction results in a noticeable spatial shift, evident in the misaligned structural lines on the ground.
    Our proposed SCPA corrects the inherent spatial drift and ensures that the model learn structurally consistent 3D geometry and robust NVS. In the error maps, blue indicates small errors, and red indicates large errors.}
    \label{fig:pose_corr_diff}
    \vspace{-3mm}
\end{figure*}

\noindent \textbf{Pose-Geometry Discrepancy.} 
During pose-free NVS training, $\mathcal{G}$ is optimized by rasterizing it onto the image-space using a set of predicted target camera parameters $(\hat{\mathcal{P}}_\text{tgt} = \{\hat{\bm{P}}_{\text{tgt},t}\}_{t=1}^T, \hat{\mathcal{K}}_\text{tgt} = \{\hat{\bm{K}}_{\text{tgt},t}\}_{t=1}^T)$ corresponding to $T$ ground-truth target views $\mathcal{I}_\text{tgt} = \{\bm{I}_{\text{tgt},t}\}_{t=1}^T$. 
This rasterization process yields the synthesized target images $\hat{\mathcal{I}}_\text{tgt} = \{\hat{\bm{I}}_{\text{tgt},t}\}_{t=1}^T$. Fig.~\ref{fig:pose_corr} illustrates two prevailing paradigms for sampling these views during training. 
The context-only approach, following \cite{jiang2025anysplat}, trivially restricts $\mathcal{I}_\text{tgt}$ to be identical to $\mathcal{I}_\text{ctx}$, leading to lack of direct supervision for novel viewpoints. Conversely, SPFSplat \cite{huang2025no} introduces a context-target strategy, sampling target views $\mathcal{I}_\text{tgt}$ that are spatially distinct from $\mathcal{I}_\text{ctx}$. This necessitates an additional forward pass to estimate $(\hat{\mathcal{P}}_\text{tgt}, \hat{\mathcal{K}}_\text{tgt})$ by feeding the concatenation of $\mathcal{I}_\text{ctx}$ and $\mathcal{I}_\text{tgt}$ into the 3DVFM.  While this disentanglement improves robustness to interpolated views since the 3D primitives $\mathcal{G}$ are derived solely from $\mathcal{I}_\text{ctx}$ and supervised by distinct target observations, it introduces a critical structural flaw. Specifically, the asymmetric information flow, where geometry extraction relies exclusively on context views while pose estimation relies on both context and target views, induces a fundamental \textit{pose-geometry discrepancy} during training. As illustrated in Fig.~\ref{fig:pose_corr_diff}, the error between the rendered view $\hat{\bm{I}}_{\text{tgt},t}$ (`Rendered View w/ $\hat{\bm{P}}_{\text{tgt},t}$') and $\bm{I}_{\text{tgt},t}$ (GT target View) is dominated by spatial misalignment rather than photometric degradation, which is the consequence of the \textit{pose-geometry discrepancy}. When supervised directly via a pixel-wise reconstruction loss, this extrinsic spatial shift yields corrupted gradients, resulting in unstable optimization.



\noindent \textbf{Self-Consistent Pose Alignment.} 
To mitigate this pose-geometry discrepancy, we propose \textit{Self-Consistent Pose Alignment} (SCPA), a self-correcting strategy that mathematically quantifies and reverses the spatial drift induced by the context-target training strategy, rather than directly applying photometric supervision on misaligned renderings. 
Based on the predicted $\mathcal{G}$ from $\mathcal{I}_\text{ctx}$ and the initial target pose estimates $\hat{\mathcal{P}}^{(1)}_\text{tgt}$ from the concatenation of ($\mathcal{I}_\text{ctx}, \mathcal{I}_\text{tgt}$), we first render an initial set of synthesized target images $\hat{\mathcal{I}}^{(1)}_\text{tgt}$. We then feed the concatenation of $\mathcal{I}_\text{ctx}$ and $\hat{\mathcal{I}}^{(1)}_\text{tgt}$ back into $f_{\phi}$ to yield a second set of pose predictions:
\begin{equation}
    \hat{\mathcal{P}}^{(2)}_\text{tgt} = f_{\phi}(\mathcal{I}_\text{ctx}, \hat{\mathcal{I}}^{(1)}_\text{tgt}).
    \label{eq:recursive_pose}
\end{equation}
Empirically, rendering a second set of images $\hat{\mathcal{I}}^{(2)}_\text{tgt}$ from $\hat{\mathcal{P}}^{(2)}_\text{tgt}$ reveals that $f_{\phi}$ exhibits a systematic, repeated drift when mapping synthesized geometry back to the pose manifold (see \textit{Suppl.} for detailed visualizations). From this observation, we propose to compute the corrected poses $\hat{\mathcal{P}}^{\text{align}}_\text{tgt} = \{\hat{\bm{P}}^{\text{align}}_{\text{tgt},t}\}_{t=1}^T$ that better align with the predicted scene geometry space by reversing the observed pose discrepancy between $\hat{\bm{P}}^{(\text{1})}_{\text{tgt}, t}$ and $\hat{\bm{P}}^{(\text{2})}_{\text{tgt}, t}$.

To approximate the aligned pose $\hat{\bm{P}}^{\text{align}}_{\text{tgt}, t}$, we compute the relative transformation between the initial prediction $\hat{\bm{P}}^{(1)}_{\text{tgt}, t}$ and the re-predicted pose $\hat{\bm{P}}^{(2)}_{\text{tgt}, t}$. We denote  this transformation as $\bm{\xi}_t$ which is a 6-vector of coordinates in the Lie algebra $\mathfrak{se}(3)$:
\begin{equation}
    \bm{\xi}_t = \log\left(\hat{\bm{P}}^{(2)}_{\text{tgt}, t} (\hat{\bm{P}}^{(1)}_{\text{tgt}, t})^{-1}\right)^{\vee},
\end{equation}
where $\log(\cdot)^{\vee}: \text{\textbf{SE}(3)} \to \mathfrak{se}(3)$ denotes the logarithm map \cite{DBLP:journals/corr/abs-2103-15980}. To perform the self-consistent pose alignment, we back-extrapolate along the manifold by negating $\bm{\xi}_t$. This negated vector is mapped back to the $\text{\textbf{SE}(3)}$ manifold via the exponential map $\exp(\cdot^{\wedge}): \mathfrak{se}(3) \to \text{\textbf{SE}(3)}$ and applied to $\hat{\bm{P}}^{(1)}_{\text{tgt}, t}$:
\begin{equation}
    \hat{\bm{P}}_{\text{tgt},t}^{\text{align}} = \exp\left((-\bm{\xi}_t)^{\wedge}\right) \hat{\bm{P}}^{(1)}_{\text{tgt}, t}.
\end{equation}
Finally, to ensure training stability and prevent degradation in cases where the re-prediction fails, we employ a minimum-error supervision strategy. We render the aligned images $\hat{\mathcal{I}}^{\text{align}}_\text{tgt}$ from $\hat{\mathcal{P}}^\text{align}_\text{tgt}$, and define the $\mathcal{L}_{\text{scpa}}$ as the minimum reconstruction error between the aligned $\hat{\bm{I}}^{\text{align}}_{\text{tgt},t}$ and initial $\hat{\bm{I}}^{(1)}_{\text{tgt},t}$ renderings:
\begin{equation}
    \mathcal{L}_{\text{scpa}} = \sum_{t=1}^{T} \min\left(\mathcal{L}_{\text{rec}}(\hat{\bm{I}}^{\text{align}}_{\text{tgt},t}, \bm{I}_{\text{tgt},t}), \mathcal{L}_{\text{rec}}(\hat{\bm{I}}^{(1)}_{\text{tgt},t}, \bm{I}_{\text{tgt},t}) \right).
\end{equation}
Following standard practices~\cite{xu2025depthsplat, chen2024mvsplat}, $\mathcal{L}_{\text{rec}}$ is defined as a weighted sum of the Mean Squared Error (MSE) and the LPIPS perceptual metric \cite{zhang2018perceptual}:
\begin{equation}
    \mathcal{L}_{\text{rec}}(\hat{\bm{I}}_{\text{tgt}}, \bm{I}_{\text{tgt}}) =  \| \hat{\bm{I}}_{\text{tgt}} - \bm{I}_{\text{tgt}} \|_2^2 + \lambda_{s} \text{LPIPS}(\hat{\bm{I}}_{\text{tgt}}, \bm{I}_{\text{tgt}}).
\end{equation}
By dynamically supervising the model with the aligned viewpoint, SCPA effectively suppresses the emergence of artifacts caused by the pose-geometry discrepancy of the context-target training strategy.

\subsection{{Rating-based Opacity Matching (ROM)}}
\label{sec:ROM}

While our SCPA successfully ensures global coordinate alignment, the predicted 3D Gaussian primitives $\mathcal{G}$ may still exhibit local geometric inconsistencies. 
These artifacts, known as `floaters,' often arise from spatial discrepancies among primitives that represent the same 3D regions but are predicted from disparate views. 
To suppress these artifacts, we introduce \textbf{Rating-based Opacity Matching (ROM)}. 
We ground this module in the paradigm of \textit{Learning from Rating-Based Feedback} \cite{arumugam2019deep, white2024rating, 10.5555/3305890.3305917}, where an agent is refined by an external oracle that provides absolute ratings of the agent's proposed states. 
As in Fig.~\ref{fig:main_architecture}, ROM consists of two stages: \textit{Teacher's Geometric Rating} and \textit{One-sided Geometric Feedback Matching}. 
In the Teacher's Geometric Rating stage, we utilize a pre-trained, lightweight sparse-view NVS model \cite{xu2025depthsplat} as an algorithmic teacher $f_{\tilde{\phi}}$. 
For any predicted Gaussian primitive $g_i$, the teacher evaluates its multi-view structural consensus to provide a geometric rating $\tilde{y}_i \in [0, 1]$. 
The primitives exhibiting high geometric error across views are assigned an `bad' rating ($\tilde{y}_i\to0$), while spatially consistent primitives are rated as `good' ($\tilde{y}_i\to1$). 
Then, in the One-sided Geometric Feedback Matching stage, we train our model $f_{\phi}$ to match its predicted ratings $y_i$ with the target ratings $\tilde{y}_i$. 
Since the `bad' primitives must be physically removed from the rendered scene, we directly formulate the predicted rating $y$ as the primitive's opacity value $\alpha$. 
Under this formulation, the feedback matching process naturally enables our model to prune artifacts by learning to drive their opacities to zero.

\noindent \textbf{Teacher's Geometric Rating.} 
To compute the geometric rating, we first quantify the local multi-view consistency of a Gaussian primitive $g_i$ predicted from the context view $\bm{I}_{\text{ctx},v}$. Specifically, we project its 3D mean $\bm{\mu}_i$ onto an adjacent context view $\bm{I}_{\text{ctx},v'}$ and measure the Euclidean distance between $\bm{\mu}_i$ and the 3D mean of the primitive sampled at that corresponding projected pixel. Let $\Pi_{v \to v'}(\bm{\mu}_i)$ denote the operation of projecting $\bm{\mu}_i$ onto the image plane of $\bm{I}_{ctx,v'}$, and then spatially sampling the 3D Gaussian mean at that projected 2D location. The scale-normalized geometric error $\epsilon_i$ for $g_i$ is formulated as:
\begin{equation}
    \epsilon_i = \frac{\big\| \bm{\mu}_i - \Pi_{v \to v'}(\bm{\mu}_i) \big\|_2}{\text{median}(\bm{D}_i) + \eta},
    \label{eq:teacher_error}
\end{equation}
where $\bm{D}_i$ represents the projected depth value, ensuring scale-invariance of $\epsilon_i$, and $\eta = 10^{-8}$ is a small constant to maintain numerical stability. Next, we extract the corresponding prediction from the teacher model $f_{\tilde{\phi}}$. By feeding the view pair $(\bm{I}_{\text{ctx},v}, \bm{I}_{\text{ctx},v'})$ into $f_{\tilde{\phi}}$, we generate the teacher's sparse-view 3D primitives $\tilde{\mathcal{G}} =  \{\tilde{g}_j\}_{j=1}^{\tilde{N}}$. Let $\tilde{g}_j$ (with 3D mean $\tilde{\bm{\mu}}_j$) represent the teacher's predicted primitive originating from the exact same pixel in $\bm{I}_{\text{ctx},v}$ as $g_i$. Following the identical protocol in Eq.~\ref{eq:teacher_error}, we compute the teacher's normalized geometric error $\tilde{\epsilon}_j$ which corresponds to our model's geometric error $\epsilon_i$.

To formulate the teacher's rating feedback, we compute the continuous excess geometric error as $E^\text{geo}_i = \max(0, \epsilon_i - \tilde{\epsilon}_j)$. Then, we generate a continuous rating $\tilde{y}_i \in (0, 1]$ for each Gaussian mean ${\bm{\mu}}_i$ that exponentially decays as our model's structural consensus degrades relative to the teacher:
\begin{equation}
    \tilde{y}_i = \exp\Big(-\lambda \cdot \text{sg}[E^\text{geo}_i])\Big)
    \label{eq:continuous_feedback}
\end{equation}
where $\lambda$ governs the decay rate (set to $5.0$), and $\text{sg}[\cdot]$ is the stop-gradient operator. In Eq.~\ref{eq:continuous_feedback}, if the geometric error of our model's prediction is equal to or lower than the teacher's one ($\epsilon_i \le \tilde{\epsilon}_j$), the target rating is set to $1$. As the excess geometric error increases, the rating smoothly approaches to $0$.

\noindent \textbf{One-sided Geometric Feedback Matching.}
The next step is to match our model's predicted rating $y$ which we define as the predicted opacity $\alpha$ with the teacher's continuous rating $\tilde{y}$. Unlike previous rating-based models \cite{luu2025enhancing, white2024rating} that enforce symmetric matching for both positive and negative feedback, we propose a \textit{one-sided matching}. Because a geometrically `bad' primitive should physically disappear from the scene, its opacity must be driven toward zero. Conversely, a `good' primitive's opacity is inherently ambiguous: it may represent a semi-transparent surface (low $\alpha$) or a solid object (high $\alpha$). Forcing valid primitives to match a strict rating of $\tilde{y}=1$ would cause severe `solidification' artifacts, destroying volumetric blending. Therefore, we treat the teacher's rating $\tilde{y}$ as a strict upper bound rather than an absolute target. To conform to the teacher's geometric feedback of `bad' primitive, our model's predicted rating (opacity) must be smaller than $\tilde{y}$. This is achieved via a pointwise margin loss $\mathcal{L}_{\text{opa}}$:
\begin{equation}
    \mathcal{L}_{\text{opa}} = \frac{1}{N} \sum_{i=1}^{N} \Big( \max(0, \alpha_i - \tilde{y}_i) \Big)^2
    \label{eq:opa_loss}
\end{equation}
Under this formulation, primitives with `good' feedback ($\tilde{y}_i = 1$) incur zero penalty for $\alpha$, yielding optimization control entirely to the photometric rendering loss $\mathcal{L}_{\text{scpa}}$. Artifacts with `bad' feedback ($\tilde{y}_i \to 0$), however, are aggressively penalized, naturally driving the network to prune floaters.

\noindent \textbf{Spatial Regularization for Geometric Consolidation.} 
For completeness, relying solely on opacity compression to remove artifacts can lead to overly sparse representations if mildly misaligned primitives are aggressively pruned. To prevent this, we complement the rating matching with a direct spatial regularization term, $\mathcal{L}_{\text{geo}}$. This loss explicitly minimizes the geometric error $\epsilon_i$ of the predicted primitives. Crucially, we clamp the maximum error input at $\tau = 2.0$. This ensures that massive errors do not produce exploding gradients that destabilize the entire 3D scene. Furthermore, the loss is weighted by the gradient-detached primitive's predicted opacity, $\text{sg}[\alpha_i]$, ensuring that the network prioritizes the spatial regularization of highly visible structures over transparent background noise. The loss is formulated as:
\begin{equation}
    \mathcal{L}_{\text{geo}} = \frac{1}{N} \sum_{i=1}^{N} \text{sg}[\alpha_i] \cdot  \min(\epsilon_i, \tau).
\end{equation}
Combining $\mathcal{L}_{\text{geo}}$ and $\mathcal{L}_{\text{opa}}$ establishes a complementary optimization framework that explicitly isolates repairable geometry from unrecoverable noise. $\mathcal{L}_{\text{geo}}$ acts as a spatial regularizer, pulling mildly deviant 3D coordinates into local multi-view consensus. However, when severe inconsistencies arise that cannot be resolved by $\mathcal{L}_{\text{geo}}$, our $\mathcal{L}_{\text{opa}}$ smoothly assumes control. It prunes these artifacts by driving their existence likelihood ($\alpha \to 0$) based on the teacher's ratings.


\subsection{Loss Functions}
The total loss $\mathcal{L}_{\text{total}}$ is defined as:
\begin{equation}
    \mathcal{L}_{\text{total}} = \mathcal{L}_{\text{scpa}} + \lambda_\text{geo} \mathcal{L}_{\text{geo}} + \lambda_\text{opa} \mathcal{L}_{\text{opa}},
\end{equation}
where $\lambda_\text{geo}$ and $\lambda_\text{opa}$ are weighting hyperparameters that balance the reconstruction fidelity against the geometric priors.

\section{Experiments}
\label{sec:experiments}

\subsection{Implementation Details}

We adopt DA3-GIANT \cite{lin2025da3} as our baseline. We follow DA3 to freeze the main encoder and the pre-trained depth/pose heads, optimizing only the Gaussian prediction head to maintain the powerful geometric priors of the foundation model. 
The Gaussian head predictions include 3D refinement for Gaussian means and other Gaussian attributes such as scales, rotations, and opacities. 
We train our model on $252 \times 252$ images for RealEstate10K (RE10K)~\cite{zhou2018stereo}, and $252 \times 448$ images for the DL3DV dataset~\cite{ling2024dl3dv}.
During each training iteration, we randomly sample 24 context views to construct the scene geometry and 8 spatially distinct target views for rendering supervision.
Training is conducted on 4 NVIDIA A100 40GB GPUs with a batch size of 1 per GPU, totaling $100,000$ iterations, while testing is performed on a single A100 40GB. Further details regarding our hyperparameters and model's weights update are provided in the \textit{Suppl}. 




\subsection{Evaluation Protocol}
We evaluate the NVS quality via PSNR, SSIM, and LPIPS~\cite{zhang2018perceptual}. We adopt the evaluation protocol established by recent pose-free NVS literature~\cite{ye2024no, huang2025no, park2026ecosplat} on the RE10K dataset. This setting stratifies testing sequences based on the visual overlap ratios between the initial and terminal frames. We utilize sequences with overlap ratios of less than $10\%$. Furthermore, we evaluate the existing methods under wide-baseline conditions on the complex DL3DV dataset by constructing challenging context-target splits. Specifically, we constrain the frame intervals between the starts and ends of the testing sequences to 90, 150, and 150 frames for the 12-, 24-, and 36-view settings, respectively. Target views are uniformly sampled across these sequences, while context views are randomly drawn from the mutually exclusive remaining frames. To analyze models' robustness, we perform the cross-dataset generalization evaluation on the ACID dataset \cite{infinite_nature_2020} as in \cite{chen2024mvsplat, huang2025no}, by utilizing the testing splits from \cite{park2026ecosplat}. Similar to prior pose-free approaches \cite{fu2024colmap, huang2025no, ye2024no,ye2026yonosplat}, we adopt the test-time pose optimization for ground truth alignment for all pose-free NVS baselines.

\subsection{NVS Performance Evaluation}

\begin{table*}[t]
\begin{center}
    \scriptsize
    \caption{Quantitative comparison of NVS performance on the RE10K dataset~\cite{zhou2018stereo} under various input-view settings. \best{Bold} indicates the best performance. OOM indicates out-of-memory inference error.}
    \vspace{-3mm}
    \resizebox{\linewidth}{!}{ 
        \setlength{\tabcolsep}{3pt}
\renewcommand{\arraystretch}{1.2}

\centering
\scriptsize
\begin{tabular}{ll | ccc | ccc | ccc}
\toprule
\multicolumn{2}{c|}{\multirow{2}{*}{Methods}} & \multicolumn{3}{c|}{12 Views} & \multicolumn{3}{c|}{24 Views} & \multicolumn{3}{c}{36 Views} \\
\cmidrule(lr){3-5} \cmidrule(lr){6-8} \cmidrule(lr){9-11}
& & PSNR$\uparrow$ & SSIM$\uparrow$ & LPIPS$\downarrow$ & PSNR$\uparrow$ & SSIM$\uparrow$ & LPIPS$\downarrow$ & PSNR$\uparrow$ & SSIM$\uparrow$ & LPIPS$\downarrow$ \\
\midrule
\multirow{3}{*}{\shortstack{w/ \\ Pose}} 
& MonoSplat~\cite{liu2025monosplat} (CVPR25) & 18.16 & 0.663 & 0.336 & 16.66 &	0.593 & 0.391 & 15.79 & 0.551 & 0.424 \\ 
& MVSplat~\cite{chen2024mvsplat} (ECCV24) & 17.98 & 	0.638 & 0.357 & 17.27 & 0.609 & 0.379  & OOM & OOM	& OOM \\
& DepthSplat~\cite{xu2025depthsplat} (CVPR25) & \second{22.56} & \second{0.793} & \second{0.200}  & 21.00 & \second{0.734} & 0.248  & 19.60 & 0.676 & 0.296 \\
\midrule
\multirow{7}{*}{\shortstack{Pose- \\ free}} 
& NoPoSplat~\cite{ye2024no} (ICLR25) & 17.15 & 0.571 & 0.437 &  17.10 & 	0.570 & 0.443  & 17.09 & 0.570 & 0.446 \\
& AnySplat~\cite{jiang2025anysplat} (SIGGRAPHAsia25) & 18.69 & 0.591 & 0.273 & 19.15 & 0.610 & 0.257 & 19.31 & 0.615 & 0.251 \\
& WorldMirror~\cite{liu2025worldmirror} (arXiv25) & 21.23	 & 0.707 & 0.267 & 21.08 & 0.701 & 0.274 & 20.98 & \second{0.699} & 0.275 \\
& SPFSplat~\cite{huang2025no} (ICCV25) & 21.57 & 0.701 & 0.254 & 21.32 & 0.694 & 0.266 & 21.17 & 0.689 & 0.273 \\
& YoNoSplat~\cite{ye2026yonosplat} (ICLR26) & 21.62 & 0.679 & 0.229 & \second{21.63} & 0.679 & \second{0.227} & \second{21.60} & 0.681 & \second{0.226} \\
& DA3~\cite{lin2025da3} (ICLR26) & 20.78 & 0.715 & 0.250 & 21.06 & 0.710 & 0.254 & 21.11 & 0.684 & 0.274 \\
& \textbf{AirSplat(Ours)} & \best{23.08} & \best{0.799} & \best{0.190} & \best{23.77} & \best{0.814} & \best{0.178} & \best{23.94} & \best{0.815} & \best{0.179} \\
\bottomrule
\end{tabular}
    }
\label{table:main_re10k}
\end{center}
\end{table*}

\begin{figure*}[t]
    \centering
    \includegraphics[width=\linewidth]{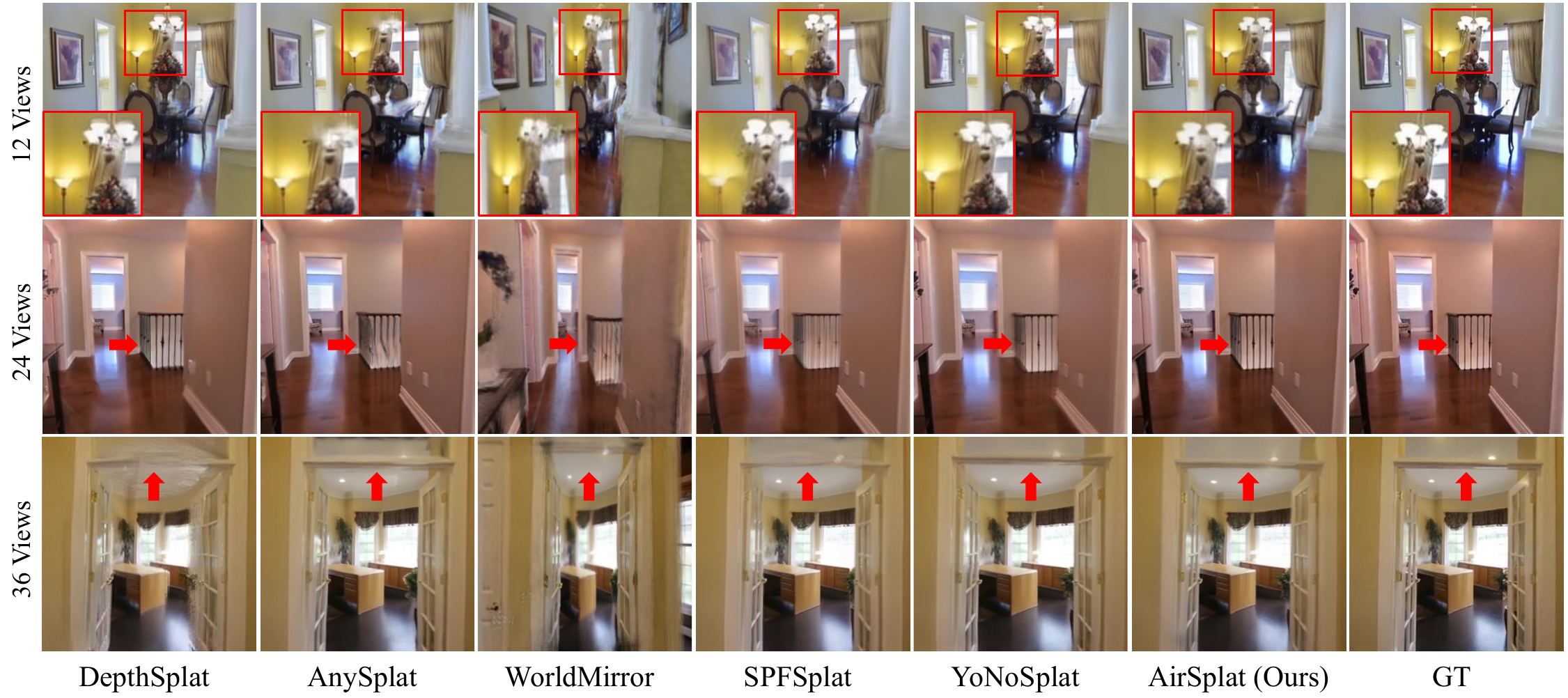}
    \vspace{-5mm}
    \caption{Qualitative comparison of NVS performance on RE10K dataset \cite{zhou2018stereo}.}
    \vspace{-3mm}
    \label{fig:qualitative_re10k_main}
\end{figure*}

\noindent \textbf{Comparison on RE10K.}
We evaluate our model on the RE10K dataset to verify its scalability and robustness in large-scale indoor and outdoor environments. 
As shown in Table~\ref{table:main_re10k}, our proposed AirSplat achieves SOTA performance among pose-free methods, demonstrating superior NVS quality.
Most notably, our method not only outperforms existing pose-free benchmarks such as YoNoSplat \cite{ye2026yonosplat} and WorldMirror \cite{liu2025worldmirror} by significant margins, but also remarkably exceeds the performance of several pose-required methods \cite{liu2025monosplat, xu2025depthsplat}. 
These results indicate that our approach effectively compensates for the absence of ground-truth camera poses by leveraging powerful geometric reasoning based on our SCPA and ROM, ultimately leading to more accurate and sharper novel view renderings compared to prior pipelines.
In Fig.~\ref{fig:qualitative_re10k_main}, our proposed AirSplat successfully preserves challenging high-frequency details, such as thin structures, while aggressively suppressing the floater artifacts and blurry regions that frequently degrade the outputs of prior methods.

\begin{table*}[t]
\begin{center}
    \scriptsize
    \caption{Quantitative comparison of NVS performance on the DL3DV dataset~\cite{ling2024dl3dv} under various input-view settings.}
    \vspace{-3mm}
    \resizebox{\linewidth}{!}{ 
        \setlength{\tabcolsep}{3pt}
\renewcommand{\arraystretch}{1.2}

\begin{tabular}{ll | ccc | ccc | ccc}
\toprule
\multicolumn{2}{c|}{\multirow{2}{*}{Methods}} & \multicolumn{3}{c|}{12 Views} & \multicolumn{3}{c|}{24 Views} & \multicolumn{3}{c}{36 Views} \\
\cmidrule(lr){3-5} \cmidrule(lr){6-8} \cmidrule(lr){9-11}
& & PSNR$\uparrow$ & SSIM$\uparrow$ & LPIPS$\downarrow$ & PSNR$\uparrow$ & SSIM$\uparrow$ & LPIPS$\downarrow$ & PSNR$\uparrow$ & SSIM$\uparrow$ & LPIPS$\downarrow$ \\
\midrule
\multirow{2}{*}{\shortstack{w/ \\ Pose}} 
& MVSplat~\cite{chen2024mvsplat} (ECCV24) & 21.55 & \second{0.729} & 0.239 & OOM & OOM & OOM & OOM & OOM & OOM \\
& DepthSplat~\cite{xu2025depthsplat} (CVPR25) & \second{22.14} & 0.725 & \second{0.221} & \second{19.87} & \second{0.695} & \second{0.274} & 18.69 & 0.643 & 0.330 \\
\midrule
\multirow{6}{*}{\shortstack{Pose- \\ free}} 
& NoPoSplat~\cite{ye2024no} (ICLR25)& 16.13 & 0.447 & 0.497 & 14.73 & 0.392 & 0.603 & 13.82 & 0.365 & 0.640 \\
& AnySplat~\cite{jiang2025anysplat} (SIGGRAPHAsia25) & 18.72 & 0.551 & 0.310 & 18.40 & 0.533 & 0.333 & 18.36 & 0.529 & 0.337 \\
& WorldMirror~\cite{liu2025worldmirror} (arXiv25) & 20.44	& 0.625	& 0.278 & 19.78 & 0.594 & 0.312 & \second{19.67} & \second{0.589} & \second{0.316} \\
& YoNoSplat~\cite{ye2026yonosplat} (ICLR26) & 17.73 & 0.481 & 0.430 & 16.77 & 0.451 & 0.490 & 16.56 & 0.446 & 0.502 \\
& DA3~\cite{lin2025da3} (ICLR26) & 20.74 & 0.691 & 0.242 & 20.51 & 0.644 & 0.274 & 20.38 & 0.642 & 0.285 \\
& \textbf{AirSplat(Ours)} & \best{22.50} & \best{0.747} & \best{0.207} & \best{22.22} & \best{0.735} & \best{0.217} & \best{22.07} & \best{0.730} & \best{0.225} \\
\bottomrule
\end{tabular}
    }
\label{table:main_dl3dv}
\end{center}
\end{table*}

\begin{figure*}[t]
    \centering
    \includegraphics[width=\linewidth]{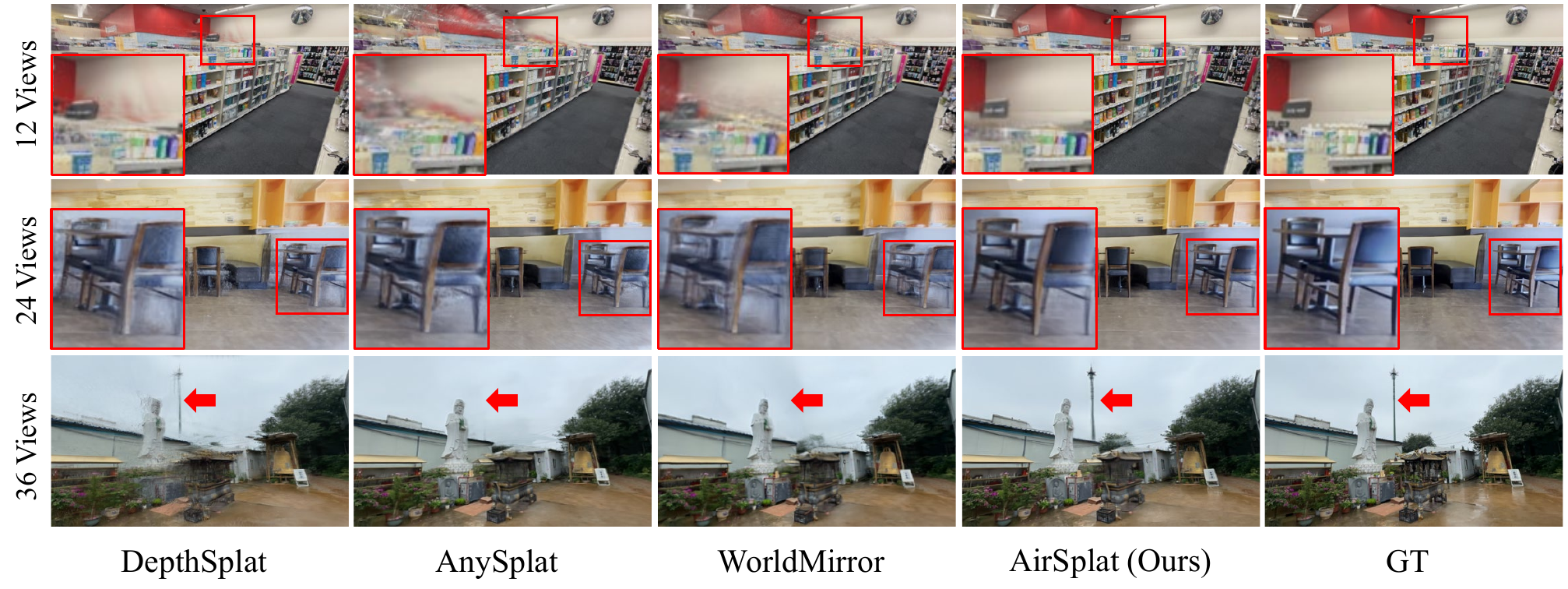}
    \vspace{-5mm}
    \caption{Qualitative comparison of NVS performance on DL3DV dataset \cite{ling2024dl3dv}.}
    \vspace{-3mm}
    \label{fig:qualitative_dl3dv_main}
\end{figure*}

\noindent \textbf{Comparison on DL3DV.}
Table \ref{table:main_dl3dv} reports the quantitative NVS performance on the DL3DV dataset compared to recent SOTA methods \cite{chen2024mvsplat, xu2025depthsplat, ye2024no, jiang2025anysplat, liu2025worldmirror, ye2026yonosplat}. 
Our AirSplat consistently achieves superior performance across all evaluation settings (12, 24, 36 views) in PSNR, SSIM, and LPIPS metrics.
Notably, in the 12-view setting, despite being a pose-free approach, our method surpasses the established SOTA pose-required method, DepthSplat \cite{xu2025depthsplat}. 
Moreover, AirSplat outperforms the baseline DA3 by a \textit{significant} margin of {+1.76 dB} in PSNR and reduces LPIPS by 14.4\%. 
While base 3D-VS-VFMs (e.g., WorldMirror~\cite{liu2025worldmirror}, DA3~\cite{lin2025da3}) are robust with increased input views, applying our proposed training strategy significantly amplifies DA3's rendering quality, yielding substantial margins in all metrics. 
Furthermore, as shown in Fig.~\ref{fig:qualitative_dl3dv_main}, our AirSplat synthesizes clean, floater-free novel views across varying input densities.
Specifically, in the third row, while other methods fail to capture thin structures, our AirSplat accurately reconstructs the sharp geometry of the pole, further validating the robust multi-view consistency and structural integrity enforced by our pipeline.

\begin{table*}[t]
\begin{center}
    \scriptsize
    \caption{Cross-dataset generalization performance under various input-view settings. We evaluate the zero-shot performance on the ACID dataset~\cite{infinite_nature_2020}.}
    \resizebox{\linewidth}{!}{ 
        \setlength{\tabcolsep}{3pt}
\renewcommand{\arraystretch}{1.2}

\centering
\scriptsize
\begin{tabular}{ll | ccc | ccc | ccc}
\toprule
\multicolumn{2}{c|}{\multirow{2}{*}{Methods}} & \multicolumn{3}{c|}{16 Views} & \multicolumn{3}{c|}{20 Views} & \multicolumn{3}{c}{24 Views} \\
\cmidrule(lr){3-5} \cmidrule(lr){6-8} \cmidrule(lr){9-11}
& & PSNR$\uparrow$ & SSIM$\uparrow$ & LPIPS$\downarrow$ & PSNR$\uparrow$ & SSIM$\uparrow$ & LPIPS$\downarrow$ & PSNR$\uparrow$ & SSIM$\uparrow$ & LPIPS$\downarrow$ \\
\midrule
\multirow{3}{*}{\shortstack{w/ \\ Pose}} 
& MonoSplat~\cite{liu2025monosplat} (CVPR25) & 17.81 & 0.642 & 0.345 & 17.48 & 0.625 & 0.361 & 17.23 & 0.611 & 0.372 \\ 
& MVSplat~\cite{chen2024mvsplat} (ECCV24) & 18.19 & 	0.502 & 0.376 & 18.12 & 0.500 &	0.379 & 18.09 & 0.500 & 0.379 \\
& DepthSplat~\cite{xu2025depthsplat} (CVPR25) & 20.41 & 0.713 & 0.268  & 19.92 & 0.694 & 0.286 & 19.78 & 0.688	 & 0.289 \\
\midrule
\multirow{7}{*}{\shortstack{Pose- \\ free}} 
& NoPoSplat~\cite{ye2024no} (ICLR25) & 22.30	 & 0.668 & 0.286 &  22.24 & 0.666 & 0.288 & 22.25 & 0.666 & 0.288  \\
& AnySplat~\cite{jiang2025anysplat} (SIGGRAPHAsia25) & 21.89 & 0.615	& 0.275 & 22.05 & 0.622 & 0.256 & 21.96	 & 0.619 & 	0.258 \\
& WorldMirror~\cite{liu2025worldmirror} (arXiv25) & 22.15 & 0.646 & 0.275 & 22.20	& 0.650 & 0.275 & 22.34	 & 0.658 & 0.273 \\
& SPFSplat~\cite{huang2025no} (ICCV25) & \second{24.58} & \second{0.725} & \second{0.218} & \second{24.49} & \second{0.722} & \second{0.221} & \second{24.40} & \second{0.720 }& 	\second{0.222} \\
& YoNoSplat~\cite{ye2026yonosplat} (ICLR26) & 22.49 & 0.641 & 0.270 & 22.48 & 0.641 & 0.271 & 22.47 & 0.642 & 0.272 \\
& DA3~\cite{lin2025da3} (ICLR26) & 22.13 & 0.687 & 0.272 & 23.21 & 0.690 & 0.262 & 23.31 & 0.694 & 0.262 \\
& \textbf{AirSplat(Ours)} & \best{25.96} & \best{0.796} & \best{0.188} & \best{26.21} & \best{0.803} & \best{0.178} & \best{26.42} & \best{0.813} & \best{0.176} \\
\bottomrule
\end{tabular}

    }
\label{table:main_acid}
\end{center}
\end{table*}

\noindent \textbf{Cross Dataset Generalization.}
Following \cite{huang2025no, park2026ecosplat}, we conduct a cross-dataset evaluation at a resolution of $252\times252$ on the ACID dataset. 
All models were evaluated using 16, 20, and 24 input views without further fine-tuning. 
As shown in Table~\ref{table:main_acid}, our method achieves the highest performance across all metrics, significantly outperforming both pose-free and pose-required baselines.
Notably, in this zero-shot setting, our model maintains a substantial lead over other pose-free NVS methods such as SPFSplat \cite{huang2025no}, YoNoSplat \cite{ye2026yonosplat}, and 3D-VS-VFMs~\cite{liu2025worldmirror, lin2025da3}.
This superior performance in unseen environments demonstrates that our AirSplat does not merely overfit to the training distribution but instead learns highly generalizable geometric representations and robust view-consistency, effectively bridging the gap between pose-free and pose-dependent NVS.


\begin{figure}[t]
    \centering
    \begin{minipage}{0.43\textwidth}
        \centering
        \includegraphics[width=\linewidth]{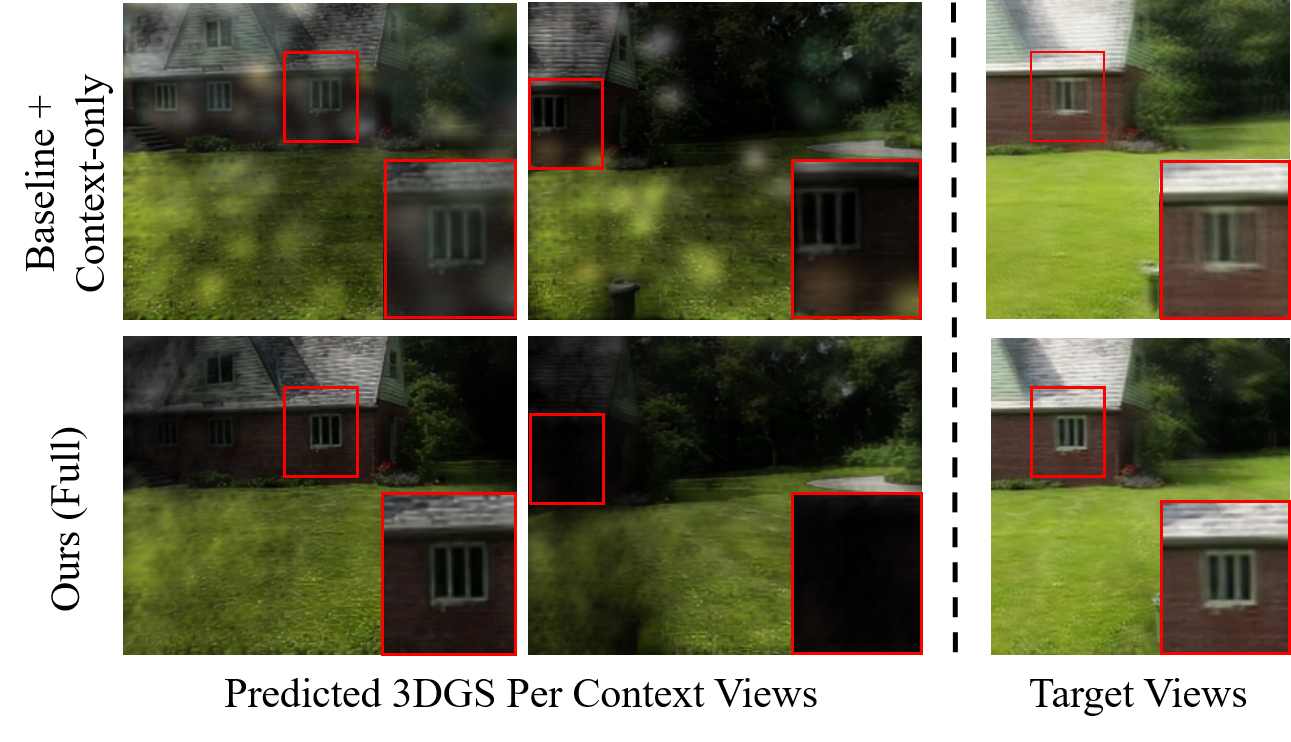} 
        \vspace{-5mm}
        \caption{Visualization of the `floaters' compression in the predicted 3DGS per context  views.}
        \vspace{-3mm}
        \label{fig:rom_ablation}
    \end{minipage}\hfill 
    \begin{minipage}{0.54\textwidth}
        \centering
        \makeatletter\def\@captype{table}\makeatother
        
        \caption{Ablation study on our SCPA and ROM. NVS performance comparison on the RE10K dataset under 12 input views.}
        \label{table:abl_main}
        
        \resizebox{\linewidth}{!}{
            \setlength{\tabcolsep}{6pt}
\renewcommand{\arraystretch}{1.2}

\begin{tabular}{l | cccc}
\toprule
Methods & PSNR$\uparrow$ & SSIM$\uparrow$ & LPIPS$\downarrow$ & GPU Mem \\
\midrule
Baseline (DA3 \cite{lin2025da3})  & 20.78 & 0.715 & 0.250 & - \\
Baseline + Context-only Training & 21.27 & 0.745 & 0.253 & 15.17 \\
Baseline + Context-Target Training & 21.63 & 0.761 & 0.241 & 19.92 \\
Baseline + Context-Target w/ SCPA Training & 22.60 & 0.776 & 0.215 & 23.45 \\
Baseline + Context-Target Training + ROM & 22.41 & 0.769 & 0.211 & 22.30\\
\textbf{Ours (Full)} & \textbf{23.08} & \textbf{0.799}	& \textbf{0.190} & 30.56 \\
\bottomrule
\end{tabular}

        }
        \vspace{-3mm}
    \end{minipage}
\end{figure}

\subsection{Ablation Study}

In Table~\ref{table:abl_main}, we conduct an ablation study to evaluate the individual contributions of our proposed components: Self-Consistent Pose Alignment (SCPA) and Rating-based Opacity Matching (ROM) on the RE10K dataset \cite{zhou2018stereo}.

\noindent \textbf{Effectiveness of SCPA.}
The baseline model, DA3~\cite{lin2025da3}, shows limited performance (PSNR: 20.78, SSIM: 0.715), which underperforms the pose-free NVS methods \cite{liu2025worldmirror, ye2024no, ye2026yonosplat}. 
As discussed in Sec.~\ref{sec:scpa}, implementing a context-only training strategy yields only a marginal improvement over the baseline (reaching 21.27 dB PSNR), inherently limited by the absence of novel target views supervision. While adopting a naive context-target sampling strategy elevates this performance to 21.63 dB, the model remains fundamentally bottlenecked by the pose-geometry discrepancy. This uncorrected spatial misalignment ultimately results in structurally degraded 3D Gaussian primitives and blurry target renderings. Integrating our SCPA results in a significant performance leap, reaching 22.60 dB (+$0.97$ dB over context-target). 
The reduction in LPIPS from 0.250 (Baseline) to 0.215 indicates that SCPA effectively restores high-frequency details by ensuring a geometrically consistent rendering path during training.

\noindent \textbf{Effectiveness of ROM.}
The integration of Rating-based Opacity Matching (ROM) independently improves the baseline to 22.41 dB PSNR. This validates the necessity of local structural consensus for high-fidelity synthesis. As illustrated in Fig.~\ref{fig:main_architecture}, ROM acts as a "geometric filter" by matching the student's predicted opacity $\alpha$ with the teacher's geometric rating $\tilde{y}$. The improvement in perceptual metrics confirms that ROM effectively prunes floaters via opacity compression. 
By explicitly penalizing spatial inconsistencies, ROM ensures that only geometrically valid structures contribute to the final rendering.
As shown in Fig.~\ref{fig:rom_ablation}, AirSplat learns from ROM to suppress inconsistent geometry of complex regions (e.g., the window structures highlighted in red boxes).
The `Baseline + Context-only' variant produces noisy `floaters' at each context view's 3DGS estimation, resulting in severe blurring artifacts.
On the other hand, our full model assigns near-zero opacity for these inconsistent primitives, resulting in sharp and high-quality reconstructions for target views. 

\noindent \textbf{Full Model.}
Our full model, AirSplat, which combines both SCPA and ROM, achieves the best performance across all metrics, yielding a 25\% improvement in perceptual scores compared to naive fine-tuning. 
Although the simultaneous integration of both modules increases the peak training memory consumption to 30.56 GB, this computational footprint is strictly confined to the offline training phase, leaving the feed-forward inference speed and memory requirements entirely unaffected.
The synergy between global pose alignment and local structural refinement allows our framework to synthesize sharp, consistent novel views from uncalibrated context images. These results validate our core hypothesis that decoupling the pose errors from geometry optimization, complemented by teacher-driven consistency feedback, is essential for robust feed-forward NVS.

\section{Limitations}
\label{sec:supp_limitations}
A primary limitation of our current framework, common among deterministic feed-forward NVS models, is its strict reliance on observed input context views. Since AirSplat does not explicitly hallucinate unseen structures, severe occlusions or entirely uncaptured areas may manifest as structural voids in the rendered novel views. To resolve these deterministic blind spots, future directions of this framework could integrate generative video diffusion priors as a post-processing module, allowing for the temporally consistent inpainting of geometrically plausible content within these occlusions. Furthermore, scaling to thousands of input context views is limited by the backbone's pixel-aligned primitives and self-attention cost, which could be addressed by pixel-unaligned primitives together with test-time training or causal attention mechanisms for multi-view feature extraction. Finally, AirSplat currently focuses on static scenes and hence may degrade on inputs with severe motion or 3D inconsistency.
\section{Conclusion}
\label{sec:conclusion}
In this paper, we presented AirSplat, a novel training framework designed for high-fidelity, pose-free novel view synthesis by effectively leveraging the geometric priors of 3D Vision Foundation Models (3DVFMs). 
Our approach identifies and addresses two fundamental challenges in existing pose-free NVS paradigms: (i) global pose-geometry discrepancy and (ii) local multi-view inconsistency.
Through the Self-Consistent Pose Alignment (SCPA), we introduce a training-time feedback loop that dynamically anchors predicted target poses to the context-derived scene geometry, effectively decoupling coordinate drift from photometric optimization.
Furthermore, we propose Rating-based Opacity Matching (ROM), which utilizes geometric feedback from a sparse-view NVS teacher model to systematically prune artifacts and floaters.
Experimental results on large-scale benchmarks, including RE10K, DL3DV and ACID, demonstrate that our AirSplat achieves state-of-the-art performance with large margins, while preserving the foundational geometry estimation performance.
\section{Acknowledgement}
\label{sec:acknowledgement}

This work was supported by Institute of Information \& communications Technology Planning \& Evaluation (IITP) grant funded by the Korean Government [Ministry of Science and ICT (Information and Communications Technology)] (Project Number: RS-2022-00144444, Project Title: Deep Learning Based Visual Representational Learning and Rendering of Static and Dynamic Scenes, 100\%).

%
%
\newpage
\clearpage
\appendix

\setcounter{page}{1}
\setcounter{linenumber}{1}

\section*{\LARGE Supplementary Material}
\label{supp}

\section{Supplementary Overview}

This supplementary document provides comprehensive implementation details, extended evaluations, and interactive visual proofs to further validate the contributions of AirSplat. Specifically, Sec.~\ref{sec:supp_implementation_details} outlines our exact hyperparameter settings and training protocols. We further provide AirSplat's adaptation of additional baseline models in Sec.~~\ref{sec:additional_baselines}. In Sec.~\ref{sec:supp_pose_geo_discrepancy}, we provide a detailed analysis of the pose-geometry discrepancy arising from the context-target training strategy. Furthermore, Sec.~\ref{sec:supp_ablation_study} validates the generalizability of our framework with additional ablation studies on the DL3DV dataset~\cite{ling2024dl3dv}, while Sec.~\ref{sec:supp_geometry_estimation} benchmarks AirSplat's visual geometry estimation against prior NVS methods. We also analyze the computational trade-offs of SCPA's training overhead in Sec.~\ref{sec:supp_training_overhead} and discuss our framework's inherent limitations in Sec.~\ref{sec:supp_limitations}. Finally, Sec.~\ref{sec:supp_experimental_results} presents extended qualitative comparisons.


\section{Implementation Details}
\label{sec:supp_implementation_details}
We employ the AdamW optimizer \cite{loshchilov2017adamw} with a learning rate of $2 \times 10^{-6}$ and a weight decay of $0.01$. 
We use a OneCycleLR \cite{smith2019super} scheduler with a warm-up period of $2,000$ steps. In Rating-based Opacity Matching (ROM), we set the error decay rate $\lambda=5.0$ and the stability constant $\eta=10^{-8}$. 
The loss weights are empirically determined as $\lambda_\text{geo}=0.1$ and $\lambda_\text{opa}=1.0$, while $\lambda_{s}$ is set to 0.1 for the LPIPS term. 
The complete training process requires approximately 4.5 days utilizing four NVIDIA A100 (40GB) GPUs.
During evaluation, we assess NVS performance using image resolutions of $252 \times 448$ for the DL3DV dataset \cite{ling2024dl3dv} and $252 \times 252$ for the RE10K dataset \cite{zhou2018stereo}.

\begin{figure*}[t]
    \centering
    \includegraphics[width=0.9\linewidth]{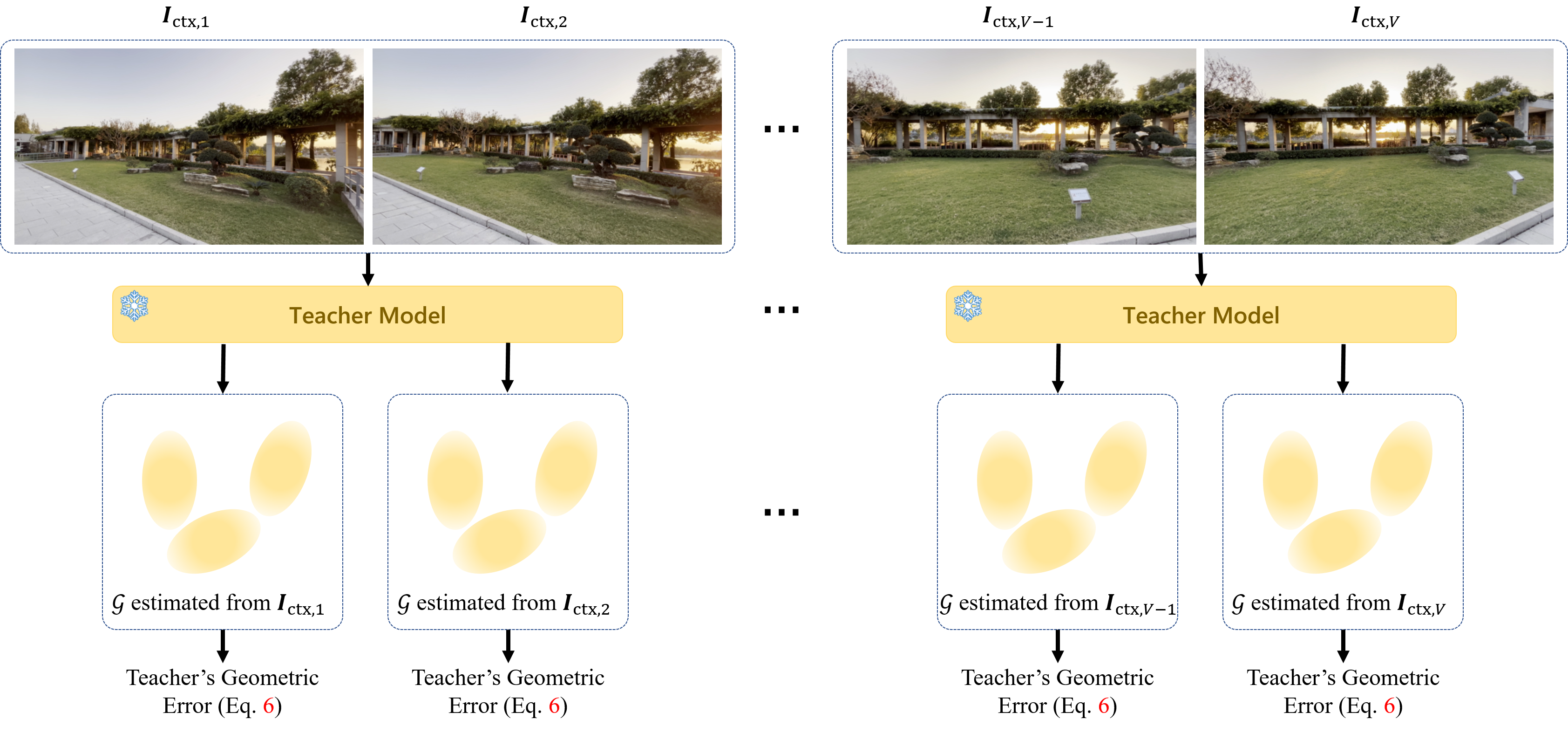}
    \caption{Our strategy for partitioning input context sequences to compute the teacher's geometric errors.}
    \label{fig:teacher_rating}
\end{figure*}

\noindent \textbf{Rating-based Opacity Matching.}
Fig.~\ref{fig:teacher_rating} further clarifies our strategy for partitioning the input sequence to compute the teacher's geometric ratings. Since we utilize a two-view feed-forward 3DGS model \cite{xu2025depthsplat} as the teacher model, the input sequence is divided into pairs of adjacent views.
It is worth noting that the teacher model does not require any calibrated or ground-truth camera poses during training; instead, it directly utilizes the predicted context poses estimated by our framework, maintaining a strictly pose-free pipeline.
The geometric errors of the pixel-aligned primitives are then computed separately for each pair using Eq.~\ref{eq:teacher_error}.
Once computed, these pairwise ratings are aggregated across the sequence to map each primitive to its corresponding opacity penalty, rigorously enforcing local multi-view consistency before dynamically compressing the student's global opacity map.

\begin{figure*}[t]
    \centering
    \includegraphics[width=1\linewidth]{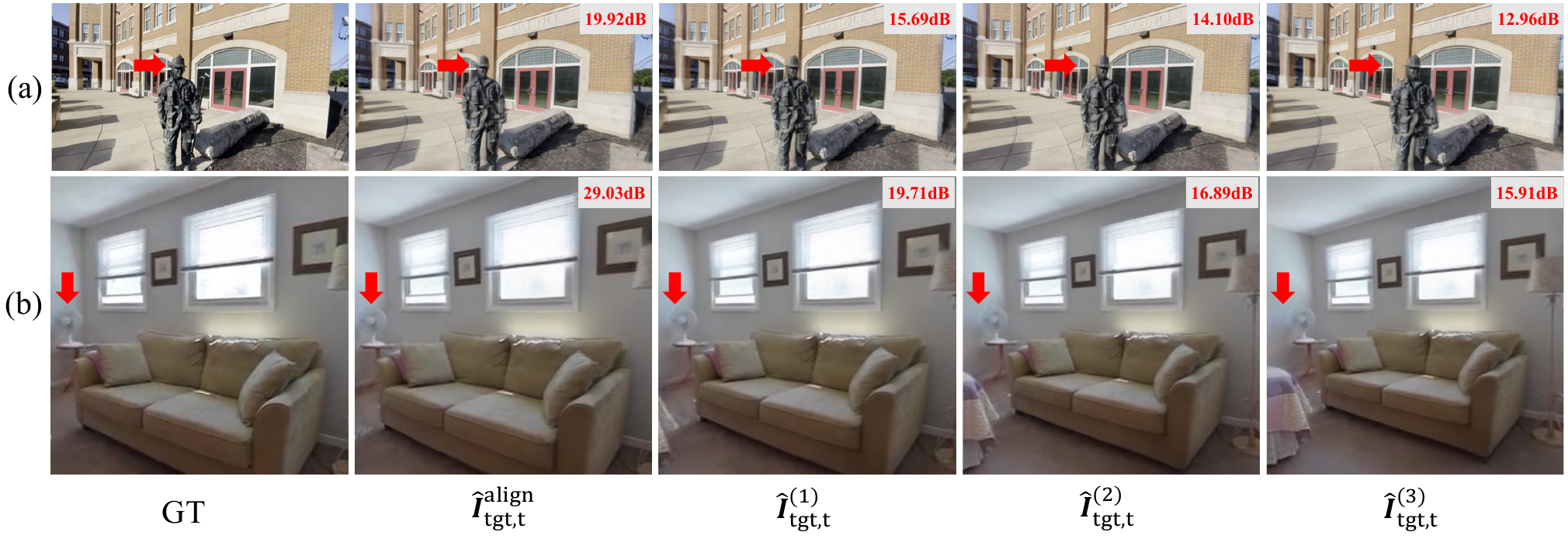}
    \caption{\textbf{Pose-Geometry Discrepancy.} To validate that this structural misalignment is not confined to a specific baseline, we visualize the consistent spatial drift across distinct models and diverse scenes: (a) DepthAnything3 \cite{lin2025da3} evaluated on the DL3DV dataset \cite{ling2024dl3dv}, and (b) SPFSplat \cite{huang2025no} on the RE10K dataset \cite{zhou2018stereo}.}
    \label{fig:supp_pose_geo_discrepancy}
\end{figure*}

\section{Additional Baselines}
\label{sec:additional_baselines}
To demonstrate that our training strategy is robust across diverse methods, we finetune two additional baselines, SPFSplat~\cite{huang2025no} and WorldMirror~\cite{liu2025worldmirror}. As shown in Table.~\ref {table:additional_baselines}  (RE10K, 24-view), AirSplat consistently improves NVS quality on both backbones. It adds a one-time training cost but yields substantial gains while preserving each backbone's native inference speed. The gain is larger for 3D-VS-VFMs, which carry strong geometric priors, than for the feed-forward 3DGS baseline (SPFSplat).

\begin{table*}[t]
\begin{center}
    \scriptsize
    \caption{Additional baselines results on RE10K dataset under 24 input views.}
    \resizebox{0.8\linewidth}{!}{ 
        \setlength{\tabcolsep}{6pt}
            \renewcommand{\arraystretch}{1.2}
            
            \begin{tabular}{l | ccc}
            \toprule
            Methods & PSNR$\uparrow$ & SSIM$\uparrow$ & LPIPS$\downarrow$ \\
            \midrule
            SPFSplat \cite{huang2025no} & 21.32 & 0.694 & 0.266 \\
            SPFSplat \textbf{+AirSplat} \cite{huang2025no} & 22.76 & 0.793 & 0.206 \\
            \midrule
            WorldMirror \cite{liu2025worldmirror} & 21.08 & 0.701 & 0.274 \\
            WorldMirror \cite{liu2025worldmirror} \textbf{+AirSplat} & 23.66 & 0.811 & 0.182 \\
            \midrule

            DA3 \cite{lin2025da3}  & 21.06 & 0.710 & 0.254 \\
            DA3 \textbf{+AirSplat} \cite{lin2025da3} & 23.77 & 0.814 & 	0.178 \\
            \bottomrule
            \end{tabular}
    }
\label{table:additional_baselines}
\end{center}
\end{table*}

\section{Pose-Geometry Discrepancy}
\label{sec:supp_pose_geo_discrepancy}

As discussed in Sec.~\ref{sec:scpa}, the asymmetric information flow within the context-target training strategy causes the predicted target poses to be misaligned with the predicted 3D Gaussian primitives. 
As shown in Fig.~\ref{fig:supp_pose_geo_discrepancy}, this pose-geometry discrepancy is a fundamental challenge, rather than an isolated issue confined to a specific baseline 3DVFM or dataset. 
To explicitly demonstrate this, we visualize the structural misalignment observed in DA3 \cite{lin2025da3} evaluated on the DL3DV dataset \cite{ling2024dl3dv}, alongside the discrepancy found in SPFSplat \cite{huang2025no} on the RE10K dataset \cite{zhou2018stereo}. 
In the figure, the first column shows the ground truth (GT) target images $\bm{I}_{\text{tgt}, t}$. 
The second column shows the aligned render $\hat{\bm{I}}^{\text{align}}_{\text{tgt}, t}$ utilizing our SCPA, achieving high spatial alignment with $\bm{I}_{\text{tgt}, t}$. 
From the third column to the last, we display the systematic and repeated drift that occurs when recursively mapping the synthesized geometry back to the pose manifold to compute the target pose $\hat{\bm{P}}^{(i)}_{\text{tgt}, t}$ and its corresponding render $\hat{\bm{I}}^{(i)}_{\text{tgt}, t}$ (as in Eq.~\ref{eq:recursive_pose}). Although the misalignment of each image in $\{\hat{\bm{I}}^{(i)}_{\text{tgt}, t}\}$ amplifies compared to $\bm{I}_{\text{tgt}, t}$, leading to a severe PSNR drop, it follows a consistent and predictable pattern. 
By leveraging this trajectory, SCPA effectively uses $\hat{\bm{P}}^{(1)}_{\text{tgt}, t}$ and $\hat{\bm{P}}^{(2)}_{\text{tgt}, t}$ to mathematically recover the aligned pose $\hat{\bm{P}}^{\text{align}}_{\text{tgt}, t}$. 
Ultimately, these consistent observations across different architectures and environments strongly validate the critical necessity of our SCPA module. 
Our SCPA successfully harnesses the robust supervisory signals of novel target views reconstruction from the context-target training strategy while strictly eliminating the detrimental gradients caused by misaligned photometric supervision.


\section{Ablation Study on DL3DV Dataset}
\label{sec:supp_ablation_study}

\begin{table*}[t]
\begin{center}
    \scriptsize
    \caption{Ablation analysis on DL3DV dataset \cite{ling2024dl3dv} under 12 input views.}
    \resizebox{0.8\linewidth}{!}{ 
        \setlength{\tabcolsep}{6pt}
\renewcommand{\arraystretch}{1.2}

\begin{tabular}{l | ccc}
\toprule
Methods & PSNR$\uparrow$ & SSIM$\uparrow$ & LPIPS$\downarrow$ \\
\midrule
Baseline  & 20.74 & 0.691 & 0.242 \\
Baseline + Context-only Training & 21.39 & 0.717 & 	0.233 \\
Baseline + Context-Target Training & 21.52 & 0.727 & 0.221 \\
Baseline + Context-Target w/ SCPA Training & 22.29 & 0.740 & 0.214 \\
Baseline + ROM & 22.11 & 0.731 & 0.218 \\
\textbf{Ours (Full)} & \textbf{22.50} & \textbf{0.747} & \textbf{0.207} \\
\bottomrule
\end{tabular}
    }
\label{table:supp_abl_dl3dv}
\end{center}
\end{table*}

To further validate the generalizability and robustness of our proposed modules across diverse and complex environments, we conduct an additional ablation study on the DL3DV dataset \cite{ling2024dl3dv}. 
Table~\ref{table:supp_abl_dl3dv} demonstrates that the performance trends strictly align with the findings presented in the main paper. 
The context-only training gains marginal improvements (reaching 21.39 dB PSNR) compared to the baseline.
While the context-target training strategy yields a slight enhancement over the context-only one (21.52 dB PSNR), it remains bottlenecked by spatial misalignment. 
Integrating our Self-Consistent Pose Alignment (SCPA) successfully mitigates this pose-geometry discrepancy, driving a substantial performance leap to 22.29 dB PSNR (+$0.77$ dB over the context-target baseline). 
Similarly, the independent integration of Rating-based Opacity Matching (ROM) effectively filters out structural inconsistencies, elevating the baseline to 22.11 dB. 
Finally, our full AirSplat framework synergizes both global pose alignment and local structural refinement to achieve the highest performance across all metrics (22.50 dB PSNR, 0.747 SSIM, and 0.207 LPIPS). 
These consistent results confirm that our framework effectively unlocks high-fidelity NVS while preserving foundational geometric priors, regardless of the dataset scale or complexity.

\section{Geometry Estimation Performance Comparison}
\label{sec:supp_geometry_estimation}
Our decision to fine-tune only the Gaussian prediction head of the 3D-VS-VFM (following the paradigm of DA3~\cite{lin2025da3}) is a deliberate design choice aimed at achieving a unified model for both robust visual geometry and high-fidelity NVS. 
To analyze the effect of fine-tuning entire 3DVFMs for NVS on visual geometry performance, we evaluate multi-view reconstruction metrics on the 7-Scenes~\cite{6619221} dataset, following $\pi^3$~\cite{wang2025pi3}. 
We compare baselines including AnySplat~\cite{jiang2025anysplat}, VGGT~\cite{wang2025vggt}, $\pi^3$~\cite{wang2025pi3}, DA3 \cite{lin2025da3}, and our AirSplat. As quantitatively demonstrated in Table~\ref{table:mv_recon}, even with training distillation, AnySplat leads to degradations of the backbone VGGT's zero-shot geometric priors. With our training strategy, AirSplat obtains improved NVS quality while maintaining the visual geometry estimation performance of DA3 \cite{lin2025da3}.

\begin{table*}[t]
\begin{center}
    \scriptsize
    \caption{Pose-free point map estimation on 7-Scenes~\cite{6619221} dataset.}
    \setlength{\tabcolsep}{7pt}
\begin{tabular}{ll cccccc}
\toprule
\multirow{2}{*}{\textbf{Method}} & \multirow{2}{*}{\textbf{View}} & \multicolumn{2}{c}{Acc. $\downarrow$} & \multicolumn{2}{c}{Comp. $\downarrow$} & \multicolumn{2}{c}{NC. $\uparrow$} \\
\cmidrule(lr){3-4} \cmidrule(lr){5-6} \cmidrule(lr){7-8}
& & Mean & Med. & Mean & Med. & Mean & Med. \\
\midrule
VGGT~\cite{wang2025vggt} & \multirow{4}{*}{\textit{sparse}} & \textbf{0.044} & \textbf{0.025} & \textbf{0.056} & \textbf{0.033} & 0.733 & 0.845 \\
AnySplat~\cite{jiang2025anysplat} & & 0.080 & 0.053 & 0.120 & 0.072 & 0.684 & 0.785 \\
$\pi^3$~\cite{wang2025pi3}& & 0.047 & 0.029 & 0.075 & 0.049 & 0.742 & 0.841 \\
DA3 \cite{lin2025da3}/AirSplat & & 0.049 & 0.034 & 0.065 & 0.046 & \textbf{0.757} & \textbf{0.866} \\
\midrule
VGGT~\cite{wang2025vggt} & \multirow{4}{*}{\textit{dense}} & 0.022 & 0.008 & 0.026 & 0.012 & 0.666 & 0.760 \\
AnySplat~\cite{jiang2025anysplat} & & 0.040 & 0.015 & 0.030 & 0.011 & 0.648 & 0.732 \\
$\pi^3$~\cite{wang2025pi3}& & \textbf{0.016} & \textbf{0.007} & \textbf{0.022} & 0.011 & \textbf{0.689} & 0.792 \\
DA3 \cite{lin2025da3}/AirSplat & & 0.018 & \textbf{0.007} & 0.023 & \textbf{0.009} & 0.688 & \textbf{0.795} \\
\bottomrule
\end{tabular}
\label{table:mv_recon}
\end{center}
\end{table*}


\section{Training Overhead}
\label{sec:supp_training_overhead}

\begin{table*}[t]
\begin{center}
    \scriptsize
    \caption{Training time analysis. We report the average time per training iteration. While our proposed modules (SCPA and ROM) introduce additional forward passes and teacher model evaluations, the overall training remains tractable.}
    \resizebox{0.6\linewidth}{!}{ 
        
\begin{tabular}{l | c c}
\toprule
Methods & Avg. Time (s) / Iter. &  \\
\midrule
Baseline + Context-Target & 2.35 & \\
Baseline + Context-Target w/ SCPA & 3.67 &  \\
Ours (Full Model) & 3.89 &  \\
\bottomrule
\end{tabular}

    }
\label{table:training_overhead}
\end{center}
\end{table*}

We analyze the computational training complexity introduced by our proposed modules. As detailed in Table~\ref{table:training_overhead}, the integration of SCPA and ROM increases the average time per training iteration by approximately 65\%. 
This expected overhead primarily stems from the supplementary forward passes required for pose correction and the teacher model evaluations for geometric rating. 
Crucially, this computational requirement is strictly confined to the training phase, imposing absolutely zero additional burden during feed-forward inference. The significant leap in state-of-the-art NVS quality and structural consistency fully justifies this trade-off of training efficiency.



\begin{figure*}[t]
    \centering
    \includegraphics[width=\linewidth]{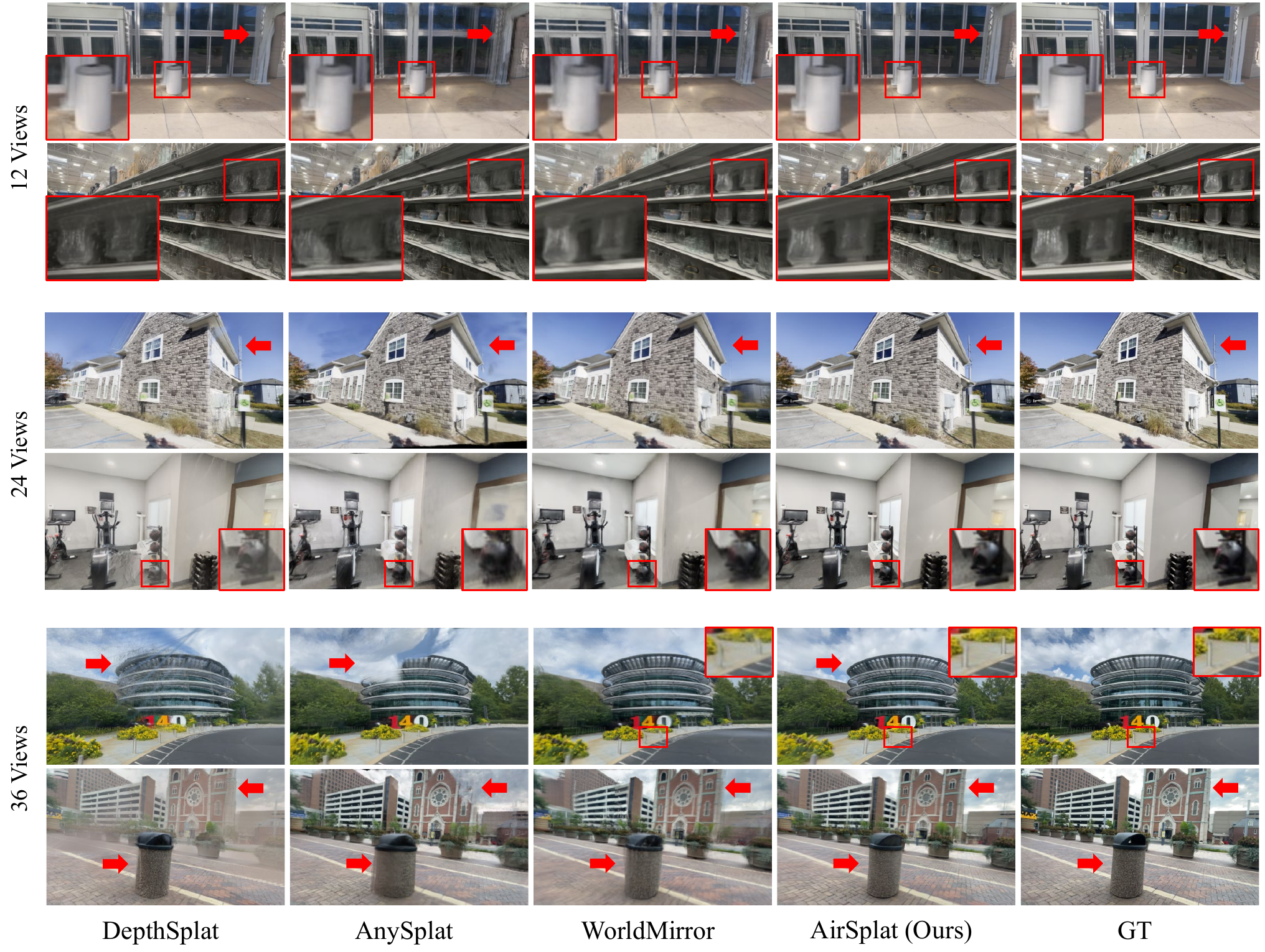}
    \caption{Qualitative comparison of NVS performance on DL3DV dataset \cite{ling2024dl3dv}.}
    \label{fig:qualitative_dl3dv_supp}
\end{figure*}

\section{Additional Qualitative Comparison}
\label{sec:supp_experimental_results}

Fig.~\ref{fig:qualitative_dl3dv_supp} provides a qualitative comparison of novel view synthesis (NVS) performance under various input-view settings on the DL3DV dataset \cite{ling2024dl3dv}. 
Compared to recent baseline methods, including DepthSplat \cite{xu2025depthsplat}, AnySplat \cite{jiang2025anysplat}, and WorldMirror \cite{liu2025worldmirror}, AirSplat consistently synthesizes significantly sharper and higher-quality renderings. 
For instance, as highlighted in the third row, AirSplat accurately reconstructs challenging high-frequency details, such as the thin structure of the pole, which are distorted or entirely missed by prior approaches. 
Furthermore, the first and last rows demonstrate our model's superior capability in preserving sharp structural boundaries. 
Notably, while baseline methods like DepthSplat, AnySplat, and WorldMirror tend to accumulate severe floater artifacts and blurring as the number of input views increases, AirSplat maintains a clean, geometrically consistent reconstruction, validating the robustness of our framework across varying input densities. 

\section{Future Works}
In future work, we plan to extend AirSplat to dynamic, feed-forward reconstruction. Two questions are central: how to make SCPA robust on non-rigid scenes, and how to distill a sparse, static feed-forward GS model into a large-scale 3D-VS-VFM that handles dynamic inputs or high-resolution inputs \cite{ngo2026dage}. We also plan to incorporate a generative prior to complete occluded regions and produce high-quality NVS from sparse observations, as explored in~\cite{kim2025exploregsexplorable3dscene, zhu2026gaussfusion, ngo2026volfill}. Finally, feed-forward reconstruction could supply physical digital twins for current robot-policy training~\cite{nguyen2026fast, nguyen2026onestep, luu2026videobasedoptimaltransportfeedbackefficient} and combine with multi-modalities~\cite{ton2025taro, pham2025mdsgen}, providing cheap, photorealistic rollouts as in~\cite{yang2026neoverse}.

\newpage
\bibliographystyle{splncs04}
\bibliography{main}

\end{document}